\documentclass[11pt,a4paper]{article}
\usepackage[hyperref]{acl2019}
\usepackage{times}
\usepackage{latexsym}
\usepackage{times}
\usepackage{helvet}
\usepackage{courier}
\usepackage{url}
\usepackage{graphicx}
\usepackage{amsmath}
\usepackage{amsfonts}
\usepackage{bm}
\usepackage{color}
\usepackage{setspace}
\usepackage{subcaption}
\usepackage{booktabs}
\usepackage{mathtools}
\usepackage{multirow}
\usepackage{float}
\usepackage{caption}
\usepackage{adjustbox}
\usepackage{cleveref}
\usepackage{makecell}
\usepackage{tabularx}
\usepackage{xcolor}
\usepackage{arydshln}

\crefname{equation}{Eq.}{Eq.}
\crefname{section}{Section}{Sections}
\crefname{subsection}{Section}{Sections}
\crefname{subsubsection}{Section}{Sections}
\crefname{figure}{Figure}{Figures}
\crefname{table}{Table}{Tables}
\crefname{subfigure}{Figure}{Figures}
\crefname{algocf}{Algorithm}{Algorithms}

\newcommand{\xhdr}[1]{\vspace{0.04in}\noindent{{\bf #1}}}
\newcommand{\xihdr}[1]{\vspace{0.02in}\noindent{{\bf #1}}}
\newcommand{\dataname}[1]{\textbf{\textit{#1}}}

\usepackage{enumitem}
\setlist{nosep,after=\vspace{0.5\baselineskip},leftmargin=12pt}
\aclfinalcopy 
\def\aclpaperid{911} 

\begin{document}
\title {Fine-Grained Spoiler Detection from Large-Scale Review Corpora}

\author{
Mengting Wan\footnotemark[1], \ Rishabh Misra\footnotemark[2], \ Ndapa Nakashole\footnotemark[1], \ Julian McAuley\footnotemark[1]\\
\footnotemark[1]~University of California, San Diego, \ \footnotemark[2]~Amazon.com, Inc\\
\{m5wan,~r1misra,~nnakashole,~jmcauley\}@ucsd.edu\\}
\maketitle

\begin{abstract}
This paper presents computational approaches for automatically detecting  critical plot twists in  reviews of media products. First, we created a large-scale book review dataset that includes fine-grained spoiler annotations at the sentence-level,  as well as book and (anonymized) user information. Second, we  carefully  analyzed this dataset, and found that:  spoiler language tends to be book-specific; spoiler distributions vary greatly across books and review authors; and spoiler sentences tend to jointly appear in the latter part of reviews. Third, inspired by these findings, we developed an end-to-end neural network architecture to detect spoiler sentences in review corpora. Quantitative and qualitative results demonstrate that the proposed method substantially outperforms existing baselines.
\end{abstract}

\section{Introduction}

`Spoilers'  on review websites can be a  concern for  consumers who want to fully experience the excitement that arises from the  pleasurable uncertainty and suspense of media consumption \cite{loewenstein1994psychology}.
Certain review websites allow reviewers to tag whether their review (or sentences in their reviews)  contain spoilers. However, we observe that in reality only a few users  utilize this feature. Thus, requiring sentence-level spoiler annotations from users is not a successful approach to comprehensive fine-grained spoiler annotation. 
One possible solution  is  crowdsourcing: whereby  consumers can report reviews that  reveal critical plot details. This is complementary to the self-reporting approach, but may have scalability issues as it is relatively difficult to engage sufficient consumers in a timely fashion.
Therefore, we seek to address the lack of completeness exhibited by  self-reporting and  crowdsourcing. We  instead  focus on developing machine learning techniques to automatically detect spoiler sentences from review documents. 

\begin{figure}[t]
  \centering
\includegraphics[width=\linewidth]{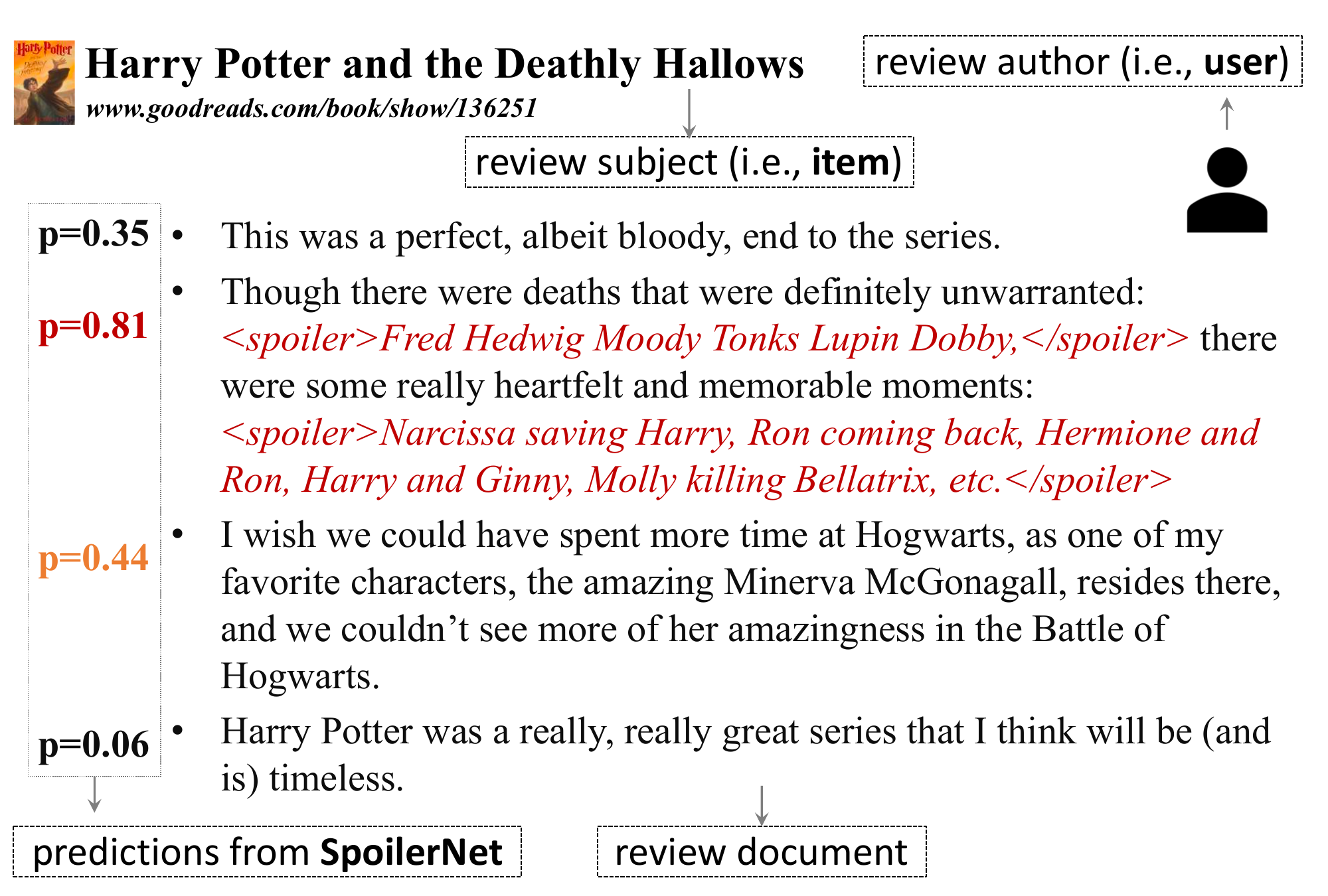}
  \caption{An example review from \dataname{Goodreads}, where spoiler tags and the predicted spoiler probabilities from \textbf{SpoilerNet} are provided.} \label{fig:example}
\end{figure}

\begin{figure*}
  \centering
  \begin{subfigure}[b]{0.195\linewidth}
\includegraphics[width=\linewidth]{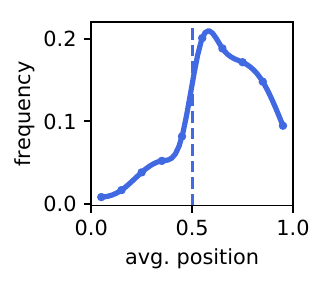}\vspace{-0.1in}
\caption{}\label{fig:sp_pos}
\end{subfigure}~
\begin{subfigure}[b]{0.195\linewidth}
\includegraphics[width=\linewidth]{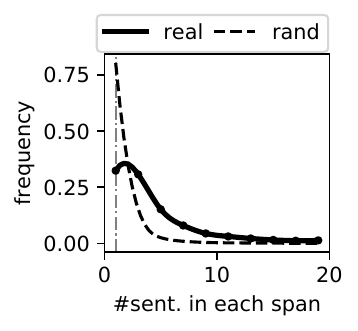}\vspace{-0.1in}
\caption{}\label{fig:sp_span}
\end{subfigure}~
\begin{subfigure}[b]{0.195\linewidth}
        \centering
        \includegraphics[width=\linewidth]{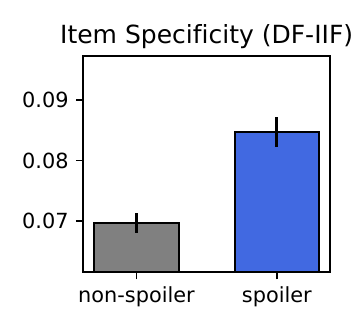}\vspace{-0.1in}
        \caption{} \label{fig:goodreads_iTfIdf}
    \end{subfigure}~
\begin{subfigure}[b]{0.195\linewidth}
        \centering
\includegraphics[width=\linewidth]{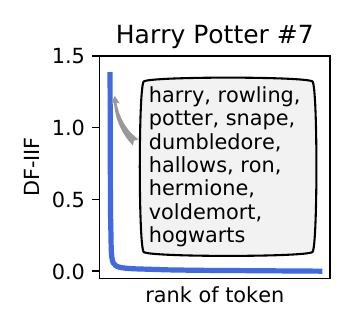}\vspace{-0.1in}
\caption{}\label{fig:top_tokens}
    \end{subfigure}~
\begin{subfigure}[b]{0.195\linewidth}
        \centering
\includegraphics[width=\linewidth]{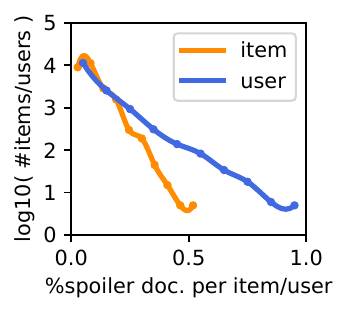}\vspace{-0.1in}
\caption{}\label{fig:dist_bias}
    \end{subfigure}
\caption{Distributions of \textbf{(a)} average spoiler sentence position; \textbf{(b)} the length of each spoiler span; \textbf{(c)} item-specificity of non-spoiler and spoiler sentences (sample means and 95\% confidence intervals); \textbf{(d)} \textbf{DF-IIF} 
of each term and top ranked item-specific terms for an example book; \textbf{(e)} the percentage of spoiler reviews per book/user.}
\end{figure*}

\xhdr{Related Work.}
Surprisingly, we find that spoiler analysis and detection is a relatively unexplored topic; previous work focuses on leveraging simple topic models \cite{guo2010finding}, or incorporating lexical features (e.g.~unigrams) \cite{boyd2013spoiler,iwai2014sentence}, frequent verbs and named entities \cite{jeon2013don}, and external meta-data of the review subjects (e.g.~genres) \cite{boyd2013spoiler} in a standard classifier such as a Support Vector Machine. Deep learning methods were first applied to this task by a recent study \cite{chang2018deep}, where the focus is modeling external genre information. Possibly due to the lack of data with complete review documents and the associated user (i.e., the review author) and item (i.e., the subject to review) ids, 
issues such as the dependency among sentences, the user/item spoiler bias, as well as the sentence semantics under different item contexts, have never been studied in this domain.

Neural network approaches have achieved great success on sentence/document classification tasks, including \textbf{CNN}-based approaches \cite{kim2014convolutional}, \textbf{RNN}-based approaches \cite{yang2016hierarchical}, and self-attention-based approaches \cite{devlin2018bert}. In this study, we cast the spoiler sentence detection task as a special sentence classification problem, but focus on modeling domain-specific language patterns.

\xhdr{Contributions.}
To address real-world, large-scale application scenarios and to facilitate the possibility of adopting modern `data-hungry' language models in this domain, we collect a new large-scale book review dataset from \textit{goodreads.com}. Spoiler tags in this dataset are self-reported by the review authors and are sentence-specific, which makes it an ideal platform for us to build supervised models. 
Motivated by the results from preliminary analysis on \dataname{Goodreads}, we propose a new model \textbf{SpoilerNet} for the spoiler sentence detection task. Using the new \dataname{Goodreads} dataset and an existing small-scale \dataname{TV Tropes} dataset \cite{boyd2013spoiler}, we demonstrate the effectiveness of the proposed techniques.

\section{The \dataname{Goodreads} Book Review Dataset}\label{sec:findings}
We scraped 1,378,033 English book reviews, across 25,475 books and 18,892 users from \textit{goodreads.com}, where each book/user has at least one associated spoiler review. These reviews include 17,672,655 sentences, 3.22\% of which are labeled as `spoiler sentences.' 
To our knowledge, this is the first dataset with fine-grained spoiler annotations at this scale. 
This dataset is available at the first author's webpage.

\xhdr{Appearance of Spoiler Sentences.} 
We first analyze the appearance of spoiler sentences in reviews by evaluating 1) the average position of spoiler sentences in a review document and 2) the average number of sentences in a spoiler span (a series of consecutive spoiler sentences). 
We present the first evaluation in \cref{fig:sp_pos}. Compared with the expected average position of randomly sampled sentences (0.5), we observe that spoiler contents tend to appear later in a review document. 
For the second evaluation, we create a benchmark distribution by randomly sampling sentences within reviews and averaging the length of each span formed by those sentences. From \cref{fig:sp_span}, compared with this random benchmark, we notice that real-world spoiler sentences tend to be `clumped' (i.e.,~more sentences in each span).

\xhdr{Item-Specificity.}
As book-specific terms such as locations or characters' names could be informative to reveal plot information \cite{jeon2013don}, we develop an effective method to identify the 
specificity of tokens regarding each item (i.e.,~each book) as follows:\footnote{$|\mathcal{D}_i|$: \#reviews associated with $i$; $|\mathcal{D}_{w,i}|$: \#reviews containing word $w$; $|\mathcal{I}_w|$: \#items containing $w$; $|\mathcal{I}|$: the total number of items. $\epsilon=1$ is a smoothing term.}
\begin{itemize}
    \item (\textbf{Popularity}) For word $w$, item $i$, we calculate the item-wise document
    frequency (\textbf{DF}) 
    as $\mathit{DF}_{w,i}=\frac{|\mathcal{D}_{w,i}|}{|\mathcal{D}_i|}$;
    \item (\textbf{Uniqueness}) For each word $w$, we calculate its inverse item frequency (\textbf{IIF}) as $\mathit{IIF}_{w}=\log\frac{|\mathcal{I}|+\epsilon}{|\mathcal{I}_w|+\epsilon}$;
    \item Then for each term $w$, item $i$, we are able to obtain the \textbf{DF-IIF} as $\mathit{DF}_{w,i}\times\mathit{IIF}_{w}$.
\end{itemize}
We show the distributions of the average \textbf{DF-IIF} values of randomly sampled non-spoiler and spoiler sentences in \cref{fig:goodreads_iTfIdf}, where we find spoilers are likely to be more book-specific.
The ranking of terms for the book \textbf{\textit{Harry Potter \#7}} is presented in \cref{fig:top_tokens}, where we find that all of the top 10 terms refer to the character/author names and important plot points. 

\xhdr{Item/User Spoilers and Self-Reporting Bias.} We further investigate the fraction of reviews containing spoiler content per item/user to analyze the spoiler appearance tendencies for each item and user (\cref{fig:dist_bias}). We notice that the distributions are highly skewed indicating significantly different spoiler tendencies across users and items. 

\xhdr{Summary of Insights.} We summarize the obtained insights as follows: 1) Spoiler sentences generally tend to appear together in the latter part of a review document, which indicates the dependency among sentences and motivates us to consider encoding such information in a spoiler detection model; 2) Item-specificity could be useful to distinguish
spoiler contents; 3) Distributions of self-reported spoiler labels are dramatically different across users and items, which motivates us to explicitly calibrate them in the model design.

\section{The Proposed Approach: SpoilerNet}
We formulate the predictive task as a binary classification problem: given a sentence $s$ in a review document, we aim to predict if it contains spoilers ($y_s=1$) or not ($y_s=0$). 

\begin{figure}[t]
  \centering
\includegraphics[width=\linewidth]{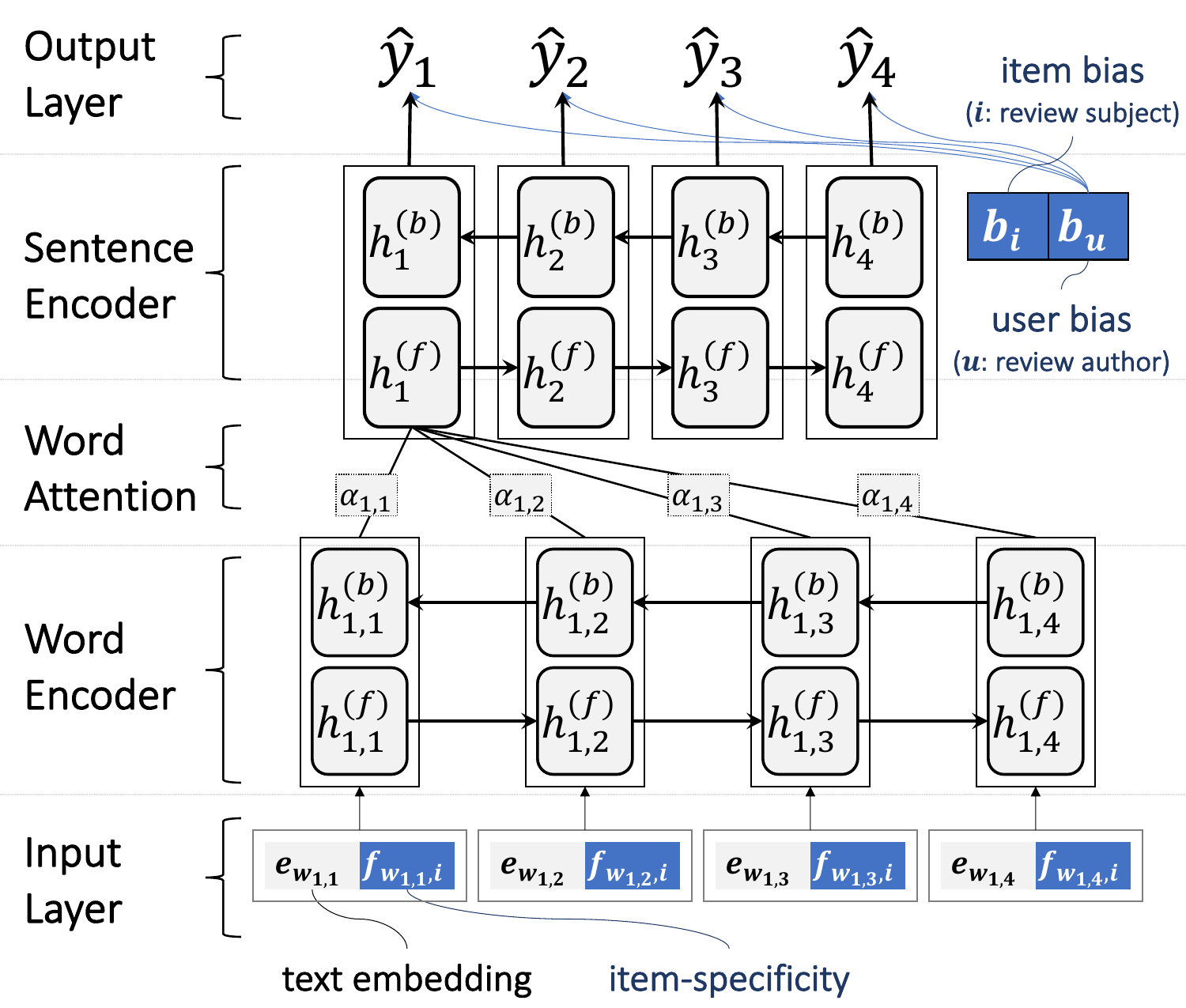}
  \caption{Model architecture of \textbf{SpoilerNet}} \label{fig:model}
\end{figure}

We introduce \textbf{SpoilerNet}, which extends the hierarchical attention network (\textbf{HAN}) \cite{yang2016hierarchical} by incorporating the above insights. We use the sentence encoder in \textbf{HAN} to model the sequential dependency among sentences. 
We incorporate the item-specificity information in the word embedding layer to enhance word representations based on different item (e.g.~book) contexts. Item and user bias terms are included in the output layer to further alleviate the disparity of spoiler distributions. \cref{fig:model} shows the overall architecture of our proposed \textbf{SpoilerNet}. We briefly describe each layer of this network as follows.

\xhdr{Input Layer.} 
For each word $w$, we introduce a $K$-dimensional text embedding $\bm{e}_w$ to represent its lexical information, which is shared across the corpus. For each word in each sentence, we calculate its corresponding \textit{item specificity features}: $\bm{f}_{w,i}=[\mathit{DF}_{w,i}, \mathit{IIF}_w, \mathit{DF}_{w,i}\times\mathit{IIF}_w]$. We expect this component could help distinguish different word semantics under different contexts (e.g., `Green' indicates a character's name with high item-specificity while it represents a color otherwise).
The concatenated vector $[\bm{e}_w; \bm{f}_{i,w}]$ is used as the input word embedding of word $w$ in sentence $s$.

\xhdr{Word Encoder, Word Attention, and Sentence Encoder.} Next we pass words through bidirectional recurrent neural networks (\textbf{bi-RNN}) with Gated Recurrent Units (\textbf{GRU}) \cite{cho-al-emnlp14}. \textbf{GRUs} accept a sequence of input embedding vectors $\bm{x}_t$ and recursively encode them into hidden states $\bm{h}_t$. 
Words are fed sequentially through a \textbf{GRU} and in reverse order through another \textbf{GRU}. Then we use the concatenation of these forward and backward hidden state vectors $\bm{h}_w=[\bm{h}_w^{(f)}; \bm{h}_w^{(b)}]$ to represent a word $w$ in a sentence $s$.

Then we introduce a word attention mechanism to focus on revelatory words (e.g., `kill', `die'), which yields
\begin{equation*}
    \begin{aligned}
        \bm{\mu}_w =& \mathit{tanh}(\bm{W}_a\bm{h}_w + \bm{b}_a), \\
        \alpha_w =& \frac{\exp(\bm{\nu}^T\bm{\mu}_w)}{\sum_{w'\in s}, \exp(\bm{\nu}^T\bm{\mu}_{w'})}, \quad
        \bm{v}_s = \sum_{w\in s} \alpha_w \bm{h}_w,
    \end{aligned}
\end{equation*}
where $\bm{W}_a$, $\bm{b}_a$ and $\bm{\nu}$ are model parameters. The weighted sums $\bm{v}_s$ are used as an input vector to represent sentence $s$ in the following sentence-level model.

Within each review, we pass the sentence input vectors $\{\bm{v}_s\}$ to another \textbf{bi-RNN} with \textbf{GRU} to encode the sequential dependency among sentences. We concatenate the resulting forward and backward hidden states to get the final representation of a sentence, i.e., $\bm{h}_s=[\bm{h}_s^{(f)}; \bm{h}_s^{(b)}]$.

\xhdr{Output Layer.}
The spoiler probability of a sentence $s$ can be calculated as
\begin{equation*}
    p_s = \sigma(\bm{w}_o^T \bm{h}_s + b_i + b_u + b).
\end{equation*}
Here for each item $i$ and each user $u$, we introduce 
learnable parameters 
$b_i, b_u$ to model the \textit{item and user biases} which can not be explained by the language model.
Then we consider minimizing the following training loss
\begin{equation*}
    \mathcal{L} = \sum~\left( y_s \log p_s + \eta(1 - y_s) \log (1 - p_s) \right), 
\end{equation*}
where $\eta$ is a hyper-parameter used to balance positive and negative labels in the training data.

\section{Experiments}
We consider the following two datasets:

\xihdr{\dataname{Goodreads}.} We use the top 20,000 frequent unigrams as our vocabulary.
    We randomly select 20\% of the reviews for testing. Among the remaining 80\%, we separate 10,000 reviews for validation and use all other reviews for training. As the distribution of spoiler labels is severely imbalanced, we decrease the weight of negative labels to 
    $\eta=0.05$, which yields best results among \{0.05, 0.1, 0.2, 0.5\} on the validation set.
    
\xihdr{\dataname{TV Tropes}} is a small-scale benchmark dataset collected from \textit{tvtropes.org} \cite{boyd2013spoiler}. This dataset contains 16,261 \textit{single-sentence} comments about 884 TV programs, which have been partitioned into 70/10/20 training/validation/test splits. All unigrams are kept in the vocabulary. As it is a balanced dataset (52.72\% of the sentences are spoilers), we set $\eta=1$.

We use the ADAM optimizer \cite{kingma2014adam} with a learning rate of 0.001, a fixed batch size (64) and dropout (0.5) in the fully connected output layer. The dimensionalities of all hidden states and the context attention vector $\bm{\nu}$ are set to $50$. Word embeddings are initialized with pretrained \textbf{fasttext} word vectors \cite{joulin2016fasttext}.

\xhdr{Baselines.} We consider the following baselines:
\begin{itemize}
    \item \textbf{SVM.~} Similar to previous studies \cite{boyd2013spoiler,jeon2013don}, we apply \textbf{SVM} with a linear kernel where counts of words are used as features.
    \item \textbf{SVM-BOW.~} Weighted averages of \textbf{fasttext} word embeddings \cite{joulin2016fasttext} 
    are used as sentence features, where the weights are \textbf{Tf-Idf}s.
    \item \textbf{CNN.~} \textbf{textCNN} \cite{kim2014convolutional} is applied where we use filter sizes 3,4, and 5, each with 50 filters.
    \item \textbf{HAN.~} The item-specificity features and the item/user bias terms are removed from \textbf{SpoilerNet}. This can be regarded as a variant of \textbf{HAN} \cite{yang2016hierarchical}.
\end{itemize}
We add the item-specificity features and the item/user bias respectively on the above baselines to evaluate their effectiveness. We remove each of the word attention module, the pre-trained word embedding initialization, and the sentence encoder from \textbf{HAN} to evaluate their performance. 

\xhdr{Evaluation.} Due to the possible \textit{subjectivity} of users' self-reported spoiler tags (i.e.,~different users may maintain different standards for various review subjects), we regard the area under the ROC curve (\textbf{AUC}) as our primary evaluation metric, i.e.,~we expect a positive spoiler sentence is ranked higher than a negative non-spoiler sentence based on $p_s$.
For \dataname{Goodreads}, we also calculate the sentence ranking \textbf{AUC} within each review document and report the average across reviews. Note this averaged document \textbf{AUC} is invariant of item/user self-reporting bias, thus the language model can be evaluated exclusively. We also report \textbf{accuracy} on \dataname{TV Tropes} so that our results can be fairly compared with existing studies \cite{boyd2013spoiler,chang2018deep}.

\begin{table}[t]
    \centering
    \setlength{\tabcolsep}{2pt}
\begin{tabular}{lllll}
\toprule
& \multicolumn{2}{c}{\dataname{Goodreads}} & \multicolumn{2}{c}{\dataname{TV Tropes}} \\
{} & AUC & AUC(d.) &    AUC &    Acc. \\
\midrule
\textbf{SVM} & 0.744 & 0.790 & 0.730 & 0.657 \\[0.5mm]
{\color{red}+} item-spec. & 0.746 {\color{red} $\uparrow$} & 0.800 {\color{red} $\uparrow$} & 0.747 {\color{red}$\uparrow$} & 0.653 {\color{blue}$\downarrow$} \\
{\color{red}+} bias       & 0.864 {\color{red} $\uparrow$} & 0.793 {\color{red} $\uparrow$} & 0.722 {\color{blue}$\downarrow$} & 0.536 {\color{blue}$\downarrow$} \\
\midrule
\textbf{SVM-BOW} & 0.692 & 0.729 & 0.756 & 0.702 \\[0.5mm]
{\color{red}+} item-spec. & 0.693 {\color{red}$\uparrow$} & 0.734 {\color{red}$\uparrow$} & 0.774 {\color{red}$\uparrow$} & 0.710 {\color{red}$\uparrow$} \\
{\color{red}+} bias       & 0.838 {\color{red}$\uparrow$} & 0.742 {\color{red}$\uparrow$} & 0.753 {\color{blue}$\downarrow$} & 0.704 {\color{red}$\uparrow$} \\
\midrule
\textbf{CNN} & 0.777 & 0.825 & 0.774 & 0.709 \\[0.5mm]
{\color{red}+} item-spec. & 0.783 {\color{red}$\uparrow$} & 0.827 {\color{red}$\uparrow$} & 0.790 {\color{red}$\uparrow$} & 0.723 {\color{red}$\uparrow$} \\
{\color{red}+} bias       & 0.812 {\color{red}$\uparrow$} & 0.822 {\color{blue}$\downarrow$} & 0.781 {\color{red}$\uparrow$} & 0.711 {\color{red}$\uparrow$} \\
\midrule
{\color{blue}-} word attn. & 0.898 {\color{blue}$\downarrow$} & 0.880 {\color{blue}$\downarrow$} & 0.760 {\color{blue}$\downarrow$} & 0.695 {\color{blue}$\downarrow$} \\ 
{\color{blue}-} word init. & 0.900 {\color{blue}$\downarrow$} & 0.880 {\color{blue}$\downarrow$} & 0.702 {\color{blue}$\downarrow$} & 0.652 {\color{blue}$\downarrow$} \\
{\color{blue}-} sent. encoder & 0.790 {\color{blue}$\downarrow$} & 0.836 {\color{blue}$\downarrow$} & - & - \\[0.5mm]
\textbf{HAN} & 0.901 & 0.884 & 0.783 & 0.720\\[0.5mm]
{\color{red}+} item-spec. & 0.906 {\color{red}$\uparrow$} & \textit{\underline{0.889}} {\color{red}$\uparrow$} & \textit{\underline{0.803}} {\color{red} $\uparrow$} & 0.733 {\color{red}$\uparrow$} \\
{\color{red}+} bias       & 0.916 {\color{red}$\uparrow$} & 0.887 {\color{red}$\uparrow$} & 0.789 {\color{red}$\uparrow$} & 0.729 {\color{red}$\uparrow$} \\[0.5mm]
\textbf{SpoilerNet} & \textit{\underline{0.919}} & \textit{\underline{0.889}} & \textit{\underline{0.803}} & \textit{\underline{0.737}} \\[0.5mm]
\bottomrule
\end{tabular}
    \caption{Spoiler sentence detection results on \dataname{Goodreads} and \dataname{TV Tropes}, where arrows indicate the performance boost ({\color{red}$\uparrow$}) or drop ({\color{blue}$\downarrow$}) compared with the base model in each group. Best results are \textit{\underline{highlighed}}. }
    \label{table:results}
\end{table}

\xhdr{Results.} Spoiler detection results are presented in \cref{table:results}, where the complete \textbf{SpoilerNet} model consistently and substantially outperform baselines on both datasets. 
The accuracy that \textbf{SpoilerNet} achieved on \dataname{TV Tropes} beats the highest one among existing methods without using external item genre information (0.723), 
but is slightly lower than the best published result (0.756) where a genre encoder is applied \cite{chang2018deep}.
We notice adding the item-specificity and user/item bias generally improves the performance of most baselines except \textbf{SVM} on \dataname{TV Tropes}. 
We find the pre-trained word embedding initialization is particularly important on \dataname{TV Tropes}. One possible reason could be that the model capacity is too large compared with this dataset so that it easily overfits without proper initialization.
Note that a substantial performance drop can be observed by removing the sentence encoder on \dataname{Goodreads}, which validates the importance of modeling sentence dependency in this task. 

\section{Error Analysis}
We provide case studies to understand the limitations of the proposed model. We show review examples for three popular books \textbf{\textit{Murder on the Orient Express}},
\textbf{\textit{The Fault in Our Stars}},
and \textbf{\textit{The Hunger Games}} 
respectively.  For each example, we provide the review text, the groudtruth spoiler tags (i.e.,~if a sentence contains spoilers or not) and the predicted spoiler probabilities from \textbf{SpoilerNet}. 

\xhdr{Distracted by Revelatory Terms.}
We find the majority of false positively predicted sentences from \textbf{SpoilerNet} can be found in this category. As shown in \cref{table:ex1}, the proposed network could be easily distracted by revelatory terms (e.g.~`murder', `killed'). This leads to a potential direction for improvement: emphasizing `difficult' negative sentences with revelatory terms during training (e.g.~by `hot' negative sampling) such that the semantic nuances can be addressed.
\begin{table}[h!]
\small
    \centering
    \setlength{\tabcolsep}{2.5pt}
    \begin{tabular}{llp{2.25in}}
    \toprule
        Prob. & Label & Review Text \\
         \midrule
0.35 & False & Language: Low (one/two usages of d*mn) \\
0.32 & False & Religion: None \\
0.39 & False & Romance: None \\
\underline{0.59} & \underline{False} & Violence: Low (It's a murder mystery! Someone is killed, but it is only ever talked about.) \\
    \bottomrule
    \end{tabular}
\caption{An example review for the book \textbf{\textit{Murder on the Orient Express}}.}\label{table:ex1}\vspace{-0.1in}
\end{table}

\xhdr{Distracted by Surrounding Sentences.} Although the model is able to capture the `coagulation'
of spoilers (i.e.,~spoiler sentences tend to appear together), it can be distracted by such a property as well. As presented in \cref{table:ex2}, the third sentence was mistakenly predicted possibly because it immediately follows a spoiler sentence and contains an item-specific revelatory term (the character name `Hazel'). This indicates the current model still needs to comprehend fine-grained sentence dependencies, so that it can decide whether to propagate or ignore the surrounding spoiler signals under different contexts.
\begin{table}[h!]
\small
    \centering
    \setlength{\tabcolsep}{2.5pt}
    \begin{tabular}{llp{2.25in}}
    \toprule
        Prob. & Label & Review Text \\
         \midrule
0.08 & False & This is not your typical teenage love story. \\ 
0.86 & True & In fact it doesn't even have a happy ending. \\
\underline{0.70} & \underline{False} & I have to say Hazel with all her pragmatism and intelligence has won me over. \\
0.43 & False & She is on the exact opposite side of the spectrum than characters like the hideous Bella Swan. \\
    \bottomrule
    \end{tabular}
\caption{An example review for the book \textbf{\textit{The Fault in Our Stars}}.}\label{table:ex2}
\end{table}

\xhdr{Inconsistent Standards of Spoiler Tags.} 
We find some self-reported labels are relatively controversial, which also verifies our suspicion regarding the subjectivity of spoiler tags. As shown in \cref{table:ex3}, the last sentence was classified as `non-spoiler' by the language model, while reported by the review author as the opposite,
probably due to its close connection to the previous spoiler sentence. Note that such an example is difficult to justify even by human annotators. This motivates us to consider spoiler detection as a ranking task instead of conventional binary classification. In this way sentences can be legitimately evaluated in the same context (e.g.~the same review document) regardless of absolute thresholds. Besides the evaluation metrics, ranking losses can also be considered in future studies.
\begin{table}[h!]
\small
    \centering
    \setlength{\tabcolsep}{2.5pt}
    \begin{tabular}{llp{2.25in}}
    \toprule
        Prob. & Label & Review Text \\
         \midrule
0.01 & False & The writing is simplistic, a little more so than befits even the 1st-person narrative of a 16-year-old.\\
0.50 & True & One of things I liked best about this is having a heroine who in addition to acting for the cameras, also has to fake her affection to someone who reciprocates far more than she feels. \\
\underline{0.15} & \underline{True} & I found it very relatable. \\
    \bottomrule
    \end{tabular}
\caption{An example review for the book \textbf{\textit{The Hunger Games}}.}\label{table:ex3}\vspace{-0.1in}
\end{table}

\section{Conclusions and Future Work}
Our new dataset, analysis of spoiler language, and positive results facilitate  several directions for future work.  For example, revising spoiler contents in a `non-spoiler' way would be an interesting language generation task. In addition to review semantics, syntax information could be incorporated in a spoiler language model. The \dataname{Goodreads} dataset may also serve as a powerful spoiler source corpus. Models and knowledge learned on this dataset could be transferred to other corpora where spoiler annotations are limited or unavailable (e.g.~detecting spoilers from tweets).

\newpage
\bibliographystyle{acl_natbib}
    \bibliography{spoiler}
    
\end{document}